\documentclass[runningheads]{llncs}
\usepackage{graphicx}
\usepackage{amsfonts}
\usepackage{hyperref}
\usepackage{multirow}
\usepackage{amsmath}

\usepackage{hyperref}
\hypersetup{
    colorlinks=true,
    linkcolor=blue,
    filecolor=magenta,      
    urlcolor=cyan,
    pdftitle={Overleaf Example},
    pdfpagemode=FullScreen,
    }

\usepackage[maxbibnames=8,
style=lncs
]{biblatex}
\usepackage{amssymb}% http://ctan.org/pkg/amssymb
\usepackage{pifont}% http://ctan.org/pkg/pifont
\begin{document}
\title{Finding-Aware Anatomical Tokens for Chest X-Ray Automated Reporting}
\titlerunning{Finding-Aware Anatomical Tokens for CXR Automated Reporting}
\author{Francesco Dalla Serra \inst{1,2} \and Chaoyang Wang \inst{1} \and
Fani Deligianni \inst{2} \and
\\Jeffrey Dalton \inst{2} \and Alison Q. O'Neil \inst{1, 3}}

\authorrunning{F. Dalla Serra et al.}

\institute{Canon Medical Research Europe, Edinburgh, United Kingdom \\ \email{francesco.dallaserra@mre.medical.canon} \and
University of Glasgow, Glasgow, United Kingdom \and
 University of Edinburgh, Edinburgh, United Kingdom}

\maketitle              % typeset the header of the contribution
\begin{abstract}

The task of radiology reporting comprises describing and interpreting the medical findings in radiographic images, including description of their location and appearance. Automated approaches to radiology reporting require the image to be encoded into a suitable token representation for input to the language model. Previous methods commonly use convolutional neural networks to encode an image into a series of \emph{image-level} feature map representations. However, the generated reports often exhibit realistic style but imperfect accuracy.
Inspired by recent works for image captioning in the general domain in which each visual token corresponds to an object detected in an image, we investigate whether using local tokens corresponding to anatomical structures can improve the quality of the generated reports.
We introduce a novel adaptation of Faster R-CNN in which \emph{finding detection} is performed for the candidate bounding boxes extracted during anatomical structure localisation. We use the resulting bounding box feature representations as our set of \emph{finding-aware} anatomical tokens. This encourages the extracted anatomical tokens to be informative about the findings they contain (required for the final task of radiology reporting). Evaluating on the MIMIC-CXR dataset \cite{Johnson2019,johnson2019mimic,goldberger2000physiobank} of chest X-Ray images, we show that task-aware anatomical tokens
give state-of-the-art performance when integrated into an automated reporting pipeline, yielding generated reports with improved clinical accuracy.

\end{abstract}

\section{Introduction}

A radiology report is a detailed text description and interpretation of the findings in a medical scan, including description of their anatomical location and appearance. For example, a Chest X-Ray (CXR) report may describe an opacity (a type of finding) in the left upper lung (the relevant anatomical location) which is diagnosed as a lung nodule (interpretation). The combination of a finding and its anatomical location influences both the diagnosis and the clinical treatment decision, since the same finding may have a different list of possible clinical diagnoses depending on the location.

\begin{figure}[t]
\centering
\includegraphics[width=\textwidth]{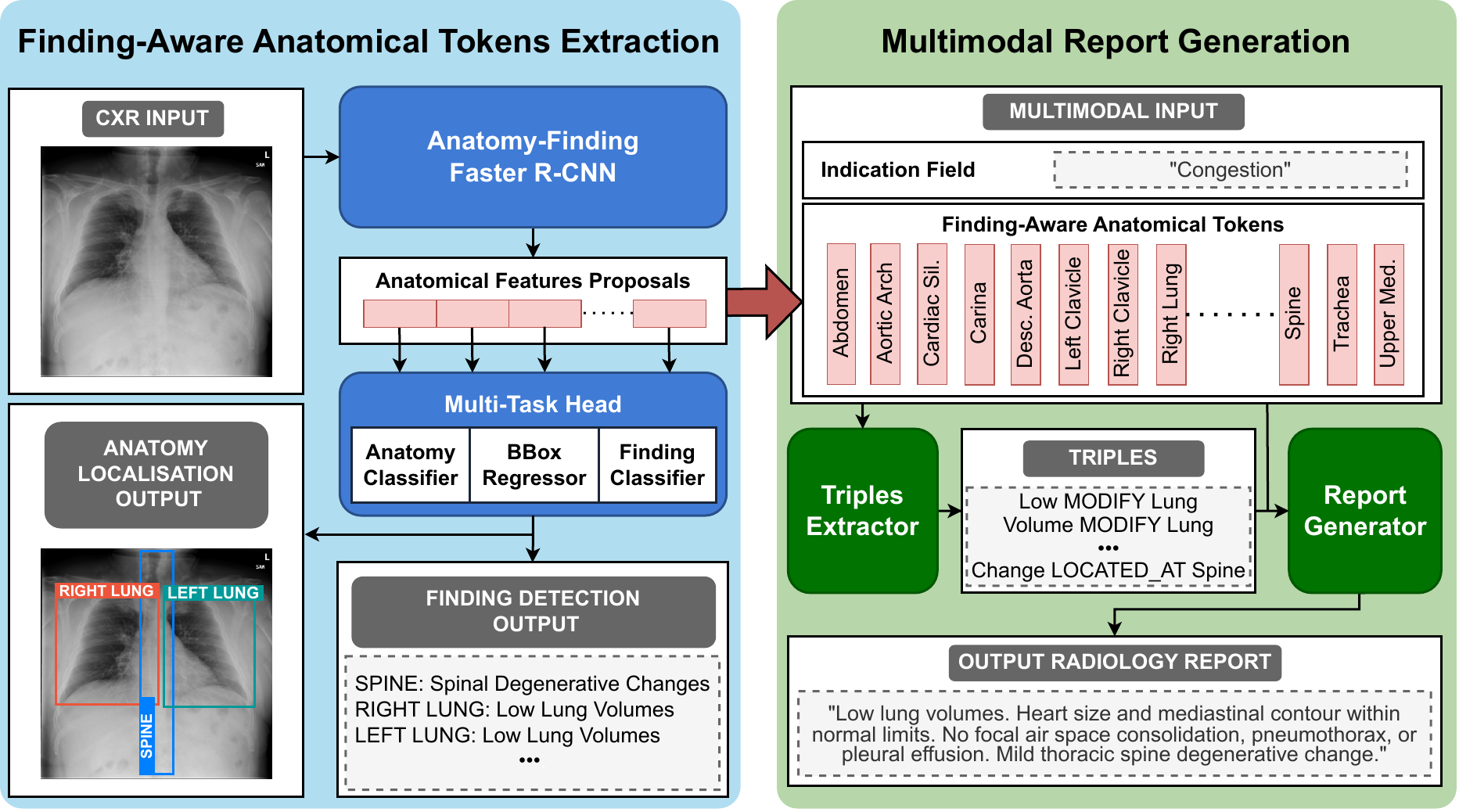}
\caption{Finding-aware anatomical tokens integrated into a multimodal CXR automated reporting pipeline. The CXR image and report are taken from the IU-Xray dataset \cite{openi}.}
\label{full_pipeline}
\end{figure}

Recent CXR automated reporting methods have adopted CNN-Transformer architectures, in which the CXR is encoded using Convolutional Neural Networks (CNNs) into global image-level features \cite{he2016deep,huang2017densely} which are input to a Transformer language model \cite{vaswani2017attention} to generate the radiology report. However, the generated reports often exhibit realistic style but imperfect accuracy, for instance hallucinating additional findings or describing abnormal regions as normal. Inspired by recent image captioning works in the general domain \cite{Anderson_2018_CVPR,li2020oscar,Zhang_2021_CVPR} in which each visual token corresponds to an object detected in the input image, we investigate whether replacing the image-level tokens with local tokens -- corresponding to anatomical structures -- can improve the clinical accuracy of the generated reports. Our contributions are to:
\begin{enumerate}
    \item Propose a novel multi-task Faster R-CNN \cite{ren2015faster} to extract \textit{finding-aware anatomical tokens} by performing finding detection on the candidate bounding boxes identified during anatomical structure localisation. We ensure these tokens convey rich information by training the model on an extensive set of anatomy regions and associated findings from the Chest ImaGenome dataset \cite{wu2021chest,goldberger2000physiobank}.
    \item Integrate the extracted finding-aware anatomical tokens as the visual input in a state-of-the-art two-stage pipeline for radiology report generation \cite{dalla2022multimodal}; this pipeline is multimodal, taking both image (CXR) and text (the corresponding text indication field) as inputs.
    \item Demonstrate the benefit of using these tokens for CXR report generation through in-depth experiments on the MIMIC-CXR dataset.
\end{enumerate}

\section{Related Works}

\paragraph{Automated Reporting}
%\label{ar_relatedworks}
Previous works on CXR automated reporting have examined the model architecture \cite{chen2020generating,chen2021cross}, the use of additional loss functions \cite{miura2021improving}, retrieval-based report generation \cite{syeda2020chest,endo2021retrieval}, and grounding report generation with structured knowledge \cite{dalla2022multimodal,yang2022knowledge}. However, no specific focus has been given to the image encoding. Inspired by recent works in image captioning in the general domain \cite{Anderson_2018_CVPR,li2020oscar,Zhang_2021_CVPR}, where each visual token corresponds to an object detected in an image, we propose to replace the image-level representations with local representations corresponding to anatomical structures detected in a CXR. To the best of our knowledge, only \cite{wang2022self,tanida2023interactive} have considered anatomical feature representations for CXR automated reporting. In \cite{wang2022self}, they extract anatomical features from an object detection model trained solely on the anatomy localisation task. In \cite{tanida2023interactive}, they train the object detector through multiple steps -- anatomy localisation, binary abnormality classification and region selection -- and feed each anatomical region individually to the language model to generate one sentence at a time. This approach makes the simplistic assumption that one anatomical region is described in exactly one report sentence. 

\paragraph{Finding Detection}
Prior works have tackled the problem of finding detection in CXR images via weakly supervised approaches \cite{yu2022anatomy,zhu2022pcan}. However, the design of these approaches does not allow the extraction of anatomy-specific vector representations, making them unsuited for our purpose. Agu et al \cite{agu2021anaxnet} proposed AnaXnet, comprising two modules trained independently: a standard Faster R-CNN trained to localise anatomical regions, and a Graph Convolutional Network (GCN) trained to classify the pathologies appearing in each anatomical region bounding box. This approach assumes that the finding information is present in the anatomical representations after the first stage of training.

\section{Methods}

We describe our method in two parts: (1) Finding-aware anatomical token extraction (Figure \ref{full_pipeline}, left) -- a custom Faster R-CNN which is trained to jointly perform \textit{anatomy localisation} and \textit{finding detection}; and (2) Multimodal report generation (Figure \ref{full_pipeline}, right) -- a two-step pipeline which is adapted to perform \textit{triples extraction} and \textit{report generation}, using the anatomical tokens extracted from the Faster R-CNN as the visual inputs for the multimodal Transformer backbone \cite{vaswani2017attention}.

\subsection{Finding-Aware Anatomical Token Extraction}
\label{sec_faster}

Let us consider $A = \{a_n\}_{n=1}^N$ as the set of anatomical regions in a CXR and $F = \{f_m\}_{m=1}^M$ the set of findings we aim to detect. We define
%$f_n = \{f_{n,m}\}_{j=1}^M$ as the findings appearing in the anatomical region $a_n$, with
$f_{n,m}\in \{0,1\}$ indicating the absence or presence of the finding $f_m$ in the anatomical region $a_n$, and $f_n = \{f_{n,m}\}_{m=1}^M$ as the set of findings in $a_n$. We define \emph{anatomy localisation} as the task of predicting the top-left and bottom-right bounding box coordinates $c=(c_{x1}, c_{y1}, c_{x2}, c_{y2})$ of the anatomical regions $A$; and \emph{finding detection} as the task of predicting the findings $f_n$ at each location $a_n$.

We frame anatomy localisation as a general object detection task, employing the Faster R-CNN framework to compute the coordinates of the bounding boxes and the anatomical labels assigned to each of them. First, the image features are extracted from the CNN backbone, composed of a ResNet-50 \cite{he2016deep} and a Feature Pyramid Network (FPN) \cite{lin2017feature}. Second, the multi-scale image features extracted from the FPN are passed to the Region Proposal Network (RPN) to generate the bounding box coordinates $c_k=(c_{k,x1}, c_{k,y1}, c_{k,x2}, c_{k,y2})$ for each proposal $k$ and to the Region of Interest (RoI) pooling layer, designed to extract the respective fixed-length vector representation $l_k\in \mathbb{R}^{1024}$. Each proposal's local features $l_k$ are then passed to a classification layer (\textit{Anatomy Classifier}) to assign the anatomical label ($a_k$) and to a bounding box regressor layer to refine the coordinates. In parallel, we insert a multi-label classification head (\textit{Findings Classifier}) -- consisting of a single fully-connected layer with sigmoid activation functions -- that classifies a set of findings for each proposal's local features (see Appendix \ref{faster_rcnn}).

During training, we use a multi-task loss comprising three terms: \textit{anatomy classification loss}, \textit{box regression loss}, and (multi-label) \textit{finding classification loss}. Formally, for each predicted bounding box, this is computed as
\begin{equation}
    \mathcal{L} = \mathcal{L}_{anatomy} + \mathcal{L}_{box} + \lambda \mathcal{L}_{finding},
\end{equation}

\noindent
where $\mathcal{L}_{anatomy}$ and $\mathcal{L}_{box}$ correspond to the anatomy classification loss and the bounding box regression loss described in \cite{girshick2015fast} and ${L}_{finding}$ is the finding classification loss that we introduce; $\lambda$ is a balancing hyper-parameter set to $\lambda=10^2$. We define
\begin{equation}
    \mathcal{L}_{finding} = - \sum_{k=1}^K\sum_{m=1}^M w_m f_{k,m} \log(p_{k,m})
\end{equation}

\noindent
a binary cross-entropy loss between the predicted probability $p_k = \{p_{k,m}\}_{m=1}^M$ of the $k$-th proposal and its associated ground truth $f_k=\{f_{k,m}\}_{m=1}^M$ (with $f_k= f_m$ if $a_k=a_m$). We class weight using $w_m=(1/\nu_m)^{\alpha}$, where $\nu_m$ is the frequency of the finding $f_m$ in the training dataset and we empirically set $\alpha$ to 0.25.

At inference time, for each CXR image, we extract the finding-aware anatomical tokens $A_{tok}=\{l_n\}_{n=1}^N$, by selecting for each anatomical region the proposal with highest anatomical classification score and taking the associated latent vector representation $l_n$. Any non-detected regions are assigned a $1024$-dimensional vector of zeros. $A_{tok}$ is provided as input to the report generation model.

\subsection{Multimodal Report Generation} \label{sec_autrep}

We adopt the multimodal knowledge-grounded approach for automated reporting on CXR images as proposed in \cite{dalla2022multimodal}. Firstly, \textit{triples extraction} is performed to extract structured information from a CXR image in the form of triples, given the indication field $Ind$ as context. Secondly, \textit{report generation} is performed to generate the radiology report from the triples with the CXR image and indication field again provided as context.

Each step is treated as a sequence-to-sequence task; for this purpose, the triples are concatenated into a single text sequence (in the order they appear in the ground truth report) separated by the special \texttt{[SEP]} token to form $Trp$, and the visual tokens are concatenated in a fixed order of anatomical regions.  Two multimodal encoder-decoder Transformers are employed as the Triples Extractor ($TE$) and Report Generator ($RG$). The overall approach is:
\begin{equation}
    \begin{aligned}
        \textrm{\textsc{Step 1}}&\quad\quad\quad Trp= TE(seg_1=A_{tok}, seg_2=Ind) \\
        \textrm{\textsc{Step 2}}&\quad\quad\quad R=  RG(seg_1=A_{tok}, seg_2= Ind \texttt{ [SEP] } Trp)
    \end{aligned}
\end{equation}

\noindent{}where $seg_1$ and $seg_2$ are the two input segments which are themselves concatenated at the input. In step 2, the indication field and the triples are merged into a single sequence of text by concatenating them, separated by the special \texttt{[SEP]} token. Similarly to \cite{devlin-etal-2019-bert}, the input to a Transformer corresponds to the sum of the textual and visual \textit{token embeddings}, the \textit{positional embeddings}---to inform about the order of the tokens---and the \textit{segment embeddings}---to discriminate between the two modalities.

\section{Experimental Setup}

\subsection{Datasets and Metrics}

We base our experiments on two open-source CXR imaging datasets, Chest ImaGenome \cite{wu2021chest,goldberger2000physiobank} and MIMIC-CXR \cite{Johnson2019,johnson2019mimic,goldberger2000physiobank}. The MIMIC-CXR dataset comprises CXR image-report pairs and is used for the target task of report generation. The Chest ImaGenome dataset is derived from MIMIC-CXR, extended with additional automatically extracted annotations for 242,072 anteroposterior and posteroanterior CXR images, which 
we use to train the finding-aware anatomical token extractor. We follow the same train/validation/test split as proposed in the Chest ImaGenome dataset. We extract the \textit{Findings} section of each report as the target text\footnote{\url{https://github.com/MIT-LCP/mimic-cxr/blob/master/txt/create_section_files.py}}.
For the textual input, we extract the \textit{Indication field} from each report.\footnote{\url{https://github.com/jacenkow/mmbt/blob/main/tools/mimic_cxr_preprocess.py}} We annotate the ground truth triples for each image-report pair following a semi-automated pipeline using RadGraph \cite{jain2021radgraph} and sciSpaCy \cite{neumann2019scispacy}, as described in \cite{dalla2022multimodal}.

To assess the quality of the generated reports, we compute Natural Language Generation (NLG) metrics: BLEU \cite{papineni2002bleu}, ROUGE \cite{lin2004rouge} and METEOR \cite{banerjee2005meteor}. We further compute Clinical Efficiency (CE) metrics by applying the CheXbert labeller \cite{smit2020combining} which extracts 14 findings to the ground truth and the generated reports, and evaluate F1, precision and recall scores. We repeat each experiment 3 times using different random seeds, reporting the average in our results.

\subsection{Implementation}

\paragraph{Finding-aware anatomical token extractor:} We adapt the Faster R-CNN implementation from \cite{li2021benchmarking}\footnote{\url{https://pytorch.org/vision/main/models/generated/torchvision.models.detection.fasterrcnn\_resnet50\_fpn\_v2.html}}, by including the finding classifier. We initialise the network with weights pre-trained on the COCO dataset \cite{lin2014microsoft}, then fine-tune it to localise 36 anatomical regions and to detect 71 findings within each region, as annotated in the Chest ImaGenome dataset (see Appendix \ref{app:anatomy-and-finding-labels}). The CXR images are resized by matching the shorter dimension to 512 pixels (maintaining the original aspect ratio) and cropping to a resolution of $512\times 512$ (random crop during training and centre crop during inference). We train the model for 25 epochs with a learning rate of $10^{-3}$, decayed every 5 epochs by a factor of 0.8. We select the model with the highest finding detection performances for the validation set, measured by computing the AUROC score for each finding at each anatomical region (see results in Appendix \ref{anatfind_results}).
\vspace{-2mm}
\paragraph{Report generator:} We implement a vanilla Transformer encoder-decoder at each step of the automated reporting pipeline. Both the encoder and the decoder consist of 3 attention layers, each composed of 8 heads and 512 hidden units. All the parameters are randomly initialised. We train step 1 for 40 epochs, with the learning rate set to $10^{-4}$ and we decay it by a factor of 0.8 every 3 epochs; and step 2 for 20 epochs, with the same learning rate as step 1. During training, we follow \cite{dalla2022multimodal} in masking out a proportion of the ground-truth triples (50\%, determined empirically), while during inference we use the triples extracted at step 1. We select the model with the highest CE-F1 score on the validation set.
\vspace{-2mm}
\paragraph{Baselines}
We benchmark against other CXR automated reporting methods: R2Gen \cite{chen2020generating}, R2GenCMN \cite{chen2021cross}, $\mathcal{M}^2$ Tr.+fact$_{\text{ENTNLI}}$ \cite{miura2021improving}, CNN+Two-Step \cite{dalla2022multimodal} and RGRG \cite{tanida2023interactive}. All these methods (except RGRG) adopt a CNN-Transformer and have shown state-of-the-art performances in report generation on the MIMIC-CXR dataset. All reported values are re-computed using the original code based on the same data split and image resolution as our method, except for \cite{tanida2023interactive} who already used this data split and image resolution, therefore we cite their reported results. We keep remaining hyperparameters as the originally reported values. 

\section{Results}

\paragraph{Overall results} In Table \ref{tab:baseline-results}, we benchmark against other state-of-the-art CXR automated reporting methods and compare with the $A_{tok}$ integrated into the full pipeline versus a simpler approach of the report generator model only, $RG$, which generates the report directly from image and indication field (omitting triples extraction). The proposed finding-aware anatomical tokens integrated with a knowledge-grounded pipeline \cite{dalla2022multimodal} generate reports with state-of-the-art fluency (NLG metrics) and clinical accuracy (CE metrics). Moreover, the superior results of our $A_{tok}$ + RG approach compared to RGRG \cite{tanida2023interactive} suggests that providing the full set of anatomical tokens together, instead of separately, gives better results. The broader visual context is indeed necessary when describing findings that span multiple regions \textit{e.g.}, assessing the position of a tube.

\begin{table}[!t]
\centering
\caption{Comparison of our proposed solution with previous approaches. TE = Triples Extractor, RG = Report Generator.}
\label{tab:baseline-results}
\resizebox{0.8\textwidth}{!}{
    \begin{tabular}{|l|cccccc|ccc|}
        \hline
        \multirow{2}{*}{\textbf{Method}} &\multicolumn{6}{c|}{\textbf{NLG}} & \multicolumn{3}{c|}{\textbf{CE}} \\
        & \textbf{BL-1} & \textbf{BL-2} & \textbf{BL-3} & \textbf{BL-4} & \textbf{MTR} & \textbf{RG-L} & \textbf{F1} & \textbf{P} & \textbf{R} \\
        \hline
        R2Gen \cite{chen2020generating} & 0.381 & 0.248 & 0.174 & 0.130 & 0.152 & 0.314 & 0.431 & 0.511 & 0.395 \\
        R2GenCMN \cite{chen2021cross} & 0.365 & 0.239 & 0.169 & 0.126 & 0.145 & 0.309 & 0.371 & 0.462 & 0.311 \\
        $\mathcal{M}^2$ Tr. + fact$_{\text{ENTNLI}}$ \cite{miura2021improving} & 0.402 & 0.261 & 0.183 & 0.136 & 0.158 & 0.300 & 0.458 & 0.540 & 0.404 \\
        ResNet-101 + $TE$ + $RG$ \cite{dalla2022multimodal} & 0.468 & 0.343 & 0.271 & 0.223 & 0.200 & 0.390 & 0.477 &	0.556 &	0.418 \\
        RGRG \cite{tanida2023interactive} & 0.400 & 0.266 & 0.187 &  0.135 & 0.168 & - & 0.461 & 0.475 & 0.447 \\
        \hline
        $A_{tok}$ + $RG$ (\textit{ours}) & 0.422 & 0.324 & 0.265	& 0.225 & 0.201	& \textbf{0.426} & 0.515 & 0.579 & 0.464
 \\
        $A_{tok}$ + $TE$ + $RG$ (\textit{ours}) & {\textbf{0.490}} & {\textbf{0.363}} & {\textbf{0.288}} & {\textbf{0.237}} & {\textbf{0.213}} & {0.406} & {\textbf{0.537}} & {\textbf{0.585}} & {\textbf{0.496}} \\
        \hline
    \end{tabular}    
}
\end{table}

\begin{table}[!t]
\centering
\caption{Comparison of different visual input representations (ResNet-101 \textit{vs.} $A_{tok}$) using different pre-training supervision (ImageNet, Findings, Anatomy and Anatomy+Findings), integrated with the TE+RG two-step pipeline.}
\label{tab:ablation-results}
\resizebox{0.8\textwidth}{!}{
    \begin{tabular}{|c|c|cccccc|ccc|}
        \hline
        \textbf{Visual } & \multirow{2}{*}{\textbf{Supervision}}
 & \multicolumn{6}{c|}{\textbf{NLG}} & \multicolumn{3}{c|}{\textbf{CE}} \\
     \textbf{Input} & & \textbf{BL-1} & \textbf{BL-2} & \textbf{BL-3} & \textbf{BL-4} & \textbf{MTR} & \textbf{RG-L} & \textbf{F1} & \textbf{P} & \textbf{R} \\
        \hline
        ResNet-101 & ImageNet & 0.468 & 0.343 & 0.271 & 0.223 & 0.200 & 0.390 & 0.477 &	0.556 &	0.418 \\
        ResNet-101 & Findings & 0.472 & 0.346 & 0.273 & 0.225 & 0.202 & 0.396 & 0.495 & 0.565 & 0.440 \\
        Naive $A_{tok}$ & Anatomy & 0.436 & 0.320 & 0.253 &  0.208 & 0.187 & 0.387 & 0.392 & 0.487 &	0.329 \\
        $A_{tok}$ & Anatomy+Findings  & {\textbf{0.490}} & {\textbf{0.363}} & {\textbf{0.288}} & {\textbf{0.237}} & {\textbf{0.213}} & {\textbf{0.406}} & {\textbf{0.537}} & {\textbf{0.585}} & {\textbf{0.496}} \\
        \hline
    \end{tabular}    
}
\end{table}

\begin{figure}[t!]
\includegraphics[width=\textwidth]{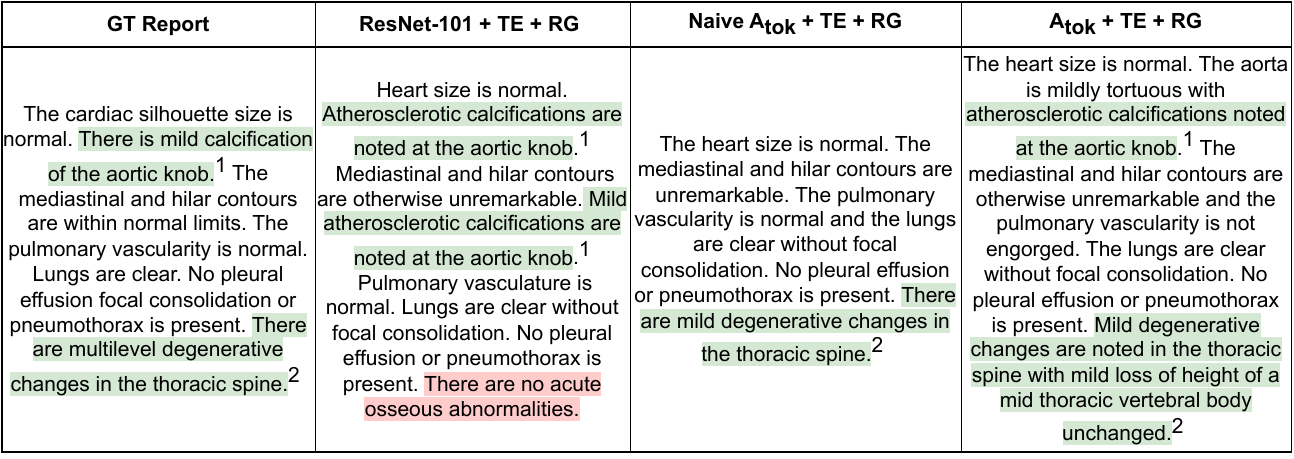}
\caption{Example report generated with different visual representations. The ground truth (GT) report (left) is followed by the generated reports using ResNet-101, naive anatomical tokens, and finding-aware anatomical tokens. Correctly detected findings are coloured green and errors are coloured red; we number corresponding descriptions.}
\label{fig:examples}
\end{figure}

\paragraph{Ablation Study} Table \ref{tab:ablation-results} shows the results of adopting different visual representations. Firstly, we use a CNN (\textbf{ResNet-101}) trained end-to-end with $TE$+$RG$ and initialised two ways: pre-trained on \textbf{ImageNet} versus pretrained on the \textbf{Findings} labels of Chest ImaGenome (details provided in Appendix \ref{cnnfind_results}). Secondly, we extract anatomical tokens (\textbf{A$_{tok}$}) with different supervision of Faster R-CNN: anatomy localisation only (\textbf{Anatomy}) or anatomy localisation + finding detection (\textbf{Anatomy+Findings}). The results show the positive effect of including supervision with finding detection either when pre-training ResNet-101 or as an additional task for Faster R-CNN. Example reports are shown in Figure \ref{fig:examples} (abbreviated) and Appendix \ref{examples} (extended).

\begin{figure}[t!]
\centering
\includegraphics[width=\textwidth]{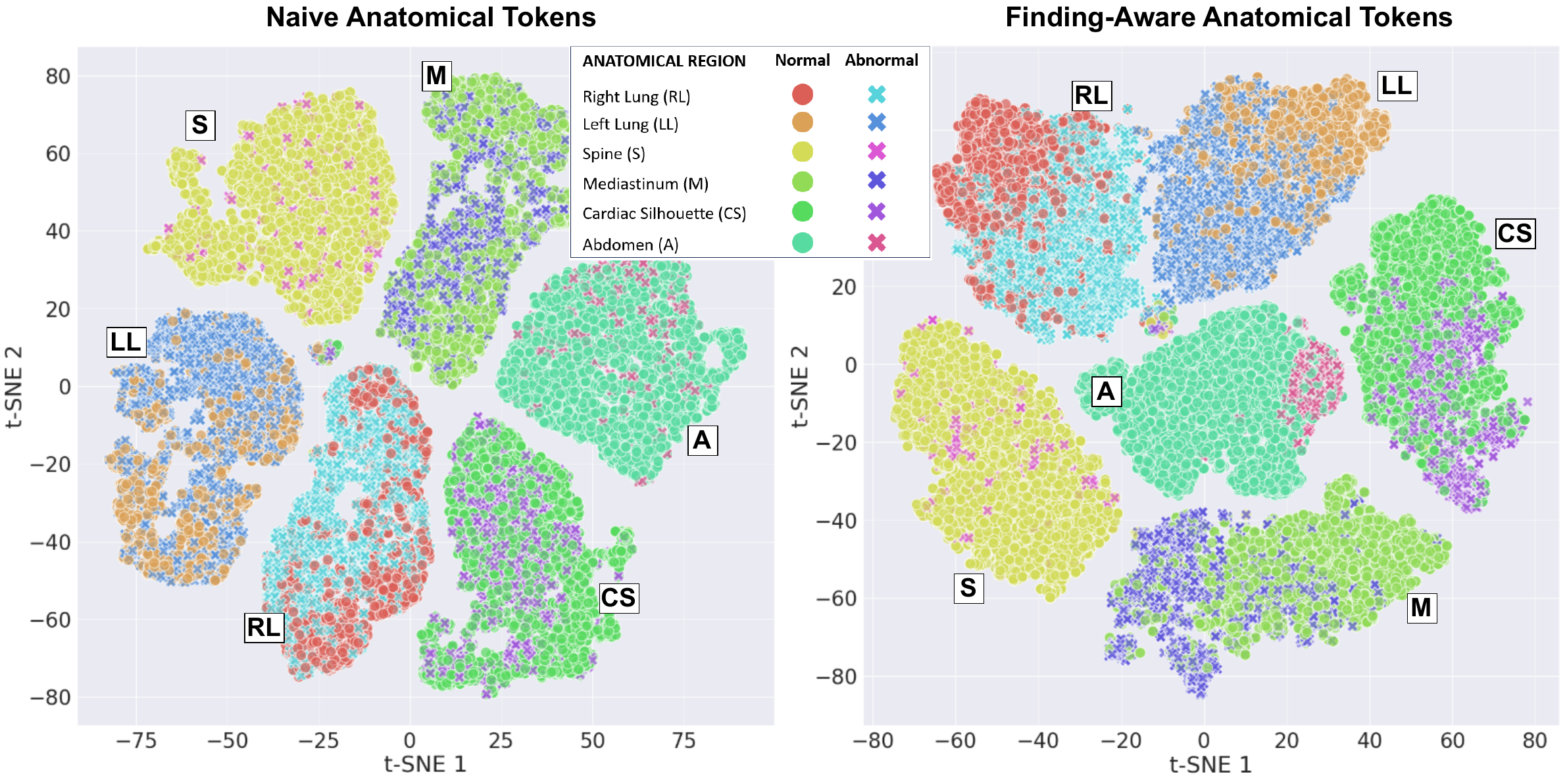}
\caption{T-SNE visualisation of normal and abnormal embeddings for a subset of visual tokens. Left: \textit{naive anatomical token} embeddings extracted from Faster R-CNN trained solely on anatomy localisation. Right: \textit{task-aware anatomical token} embeddings extracted from Faster R-CNN trained also on the finding detection task.} \label{tsne}
\end{figure}

\paragraph{Anatomical Embedding Distributions} In Figure \ref{tsne}, we visualise the impact of the finding detection task on the extracted anatomical tokens. To generate these plots, for 3000 randomly selected test set scans, we first perform principle component analysis \cite{Jolliffe} for dimensionality reduction of the token embeddings (from $\mathbb{R}^{1024}$ to $\mathbb{R}^{50}$), then use t-distributed stochastic neighbour embedding (t-SNE) \cite{van2008visualizing}, colour coding the extracted embeddings by their anatomical region and additionally categorising  as \textit{normal} or \textit{abnormal} (a token is considered abnormal if at least one of the 71 findings is positively labeled). For most anatomical regions, the normal and abnormal groups are better separated by the finding-aware tokens, suggesting these tokens successfully transmit information about findings. We also compute the mean distance between normal and abnormal clusters using Fréchet Distance (mFD) \cite{dowson1982frechet}, measuring mFD=8.80 (naive anatomical tokens) and mFD=78.67 (finding-aware anatomical tokens).

\section{Conclusion}

This work explores how to extract and integrate anatomical visual representations with language models, targeting the task of automated radiology reporting. We propose a novel multi-task Faster R-CNN adaptation that performs finding detection jointly with anatomy localisation, to extract \emph{finding-aware anatomical tokens}. We then integrate these tokens as the visual input for a multimodal image+text report generation pipeline, showing that finding-aware anatomical tokens improve the fluency (NLG metrics) and clinical accuracy (CE metrics) of the generated reports, giving state-of-the-art results.

\printbibliography

@inproceedings{girshick2015fast,
  title={Fast R-CNN},
  author={Girshick, Ross},
  booktitle={ICCV},
  pages={1440--1448},
  year={2015}
}

@inproceedings{lin2017feature,
  title={Feature pyramid networks for object detection},
  author={Lin, Tsung-Yi and Doll{\'a}r, Piotr and Girshick, Ross and He, Kaiming and Hariharan, Bharath and Belongie, Serge},
  booktitle={CVPR},
  pages={2117--2125},
  year={2017}
}

@inproceedings{dalla2022multimodal,
  title={Multimodal Generation of Radiology Reports using Knowledge-Grounded Extraction of Entities and Relations},
  author={Dalla Serra, Francesco and Clackett, William and MacKinnon, Hamish and Wang, Chaoyang and Deligianni, Fani and Dalton, Jeff and O’Neil, Alison Q},
  booktitle={AACL-IJCNLP},
  pages={615--624},
  year={2022}
}

@inproceedings{chen2020generating,
  title={Generating Radiology Reports via Memory-driven Transformer},
  author={Chen, Zhihong and Song, Yan and Chang, Tsung-Hui and Wan, Xiang},
  booktitle={EMNLP},
  pages={1439--1449},
  year={2020}
}

@inproceedings{chen2021cross,
  title={Cross-modal Memory Networks for Radiology Report Generation},
  author={Chen, Zhihong and Shen, Yaling and Song, Yan and Wan, Xiang},
  booktitle={ACL-IJCNLP},
  pages={5904--5914},
  year={2021}
}

@InProceedings{Anderson_2018_CVPR,
author = {Anderson, Peter and He, Xiaodong and Buehler, Chris and Teney, Damien and Johnson, Mark and Gould, Stephen and Zhang, Lei},
title = {Bottom-Up and Top-Down Attention for Image Captioning and Visual Question Answering},
booktitle = {CVPR},
year = {2018}
}

@inproceedings{li2020oscar,
  title={Oscar: Object-Semantics Aligned Pre-training for Vision-Language Tasks},
  author={Li, Xiujun and Yin, Xi and Li, Chunyuan and Zhang, Pengchuan and Hu, Xiaowei and Zhang, Lei and Wang, Lijuan and Hu, Houdong and Dong, Li and Wei, Furu and others},
  booktitle={ECCV},
  pages={121--137},
  year={2020},
  organization={Springer}
}

@InProceedings{Zhang_2021_CVPR,
    author    = {Zhang, Pengchuan and Li, Xiujun and Hu, Xiaowei and Yang, Jianwei and Zhang, Lei and Wang, Lijuan and Choi, Yejin and Gao, Jianfeng},
    title     = {VinVL: Revisiting Visual Representations in Vision-Language Models},
    booktitle = {CVPR},
    year      = {2021},
    pages     = {5579-5588}
}

@article{ren2015faster,
  title={Faster R-CNN: Towards real-time object detection with region proposal networks},
  author={Ren, Shaoqing and He, Kaiming and Girshick, Ross and Sun, Jian},
  journal={NIPS},
  volume={28},
  year={2015}
}

@inproceedings{agu2021anaxnet,
  title={AnaXNet: Anatomy Aware Multi-label Finding Classification in Chest X-Ray},
  author={Agu, Nkechinyere N and Wu, Joy T and Chao, Hanqing and Lourentzou, Ismini and Sharma, Arjun and Moradi, Mehdi and Yan, Pingkun and Hendler, James},
  booktitle={MICCAI},
  year={2021},
  organization={Springer}
}

@inproceedings{wu2021chest,
  title={Chest ImaGenome Dataset for Clinical Reasoning},
  author={Wu, Joy T and Agu, Nkechinyere Nneka and Lourentzou, Ismini and Sharma, Arjun and Paguio, Joseph Alexander and Yao, Jasper Seth and Dee, Edward Christopher and Mitchell, William G and Kashyap, Satyananda and Giovannini, Andrea and others},
  booktitle={NeurIPS: Datasets and Benchmarks Track (Round 2)},
  year={2021}
}

@article{johnson2019mimic,
  title={{MIMIC-CXR-JPG}, a large publicly available database of labeled chest radiographs},
  author={Johnson, Alistair EW and Pollard, Tom J and Greenbaum, Nathaniel R and Lungren, Matthew P and Deng, Chih-ying and Peng, Yifan and Lu, Zhiyong and Mark, Roger G and Berkowitz, Seth J and Horng, Steven},
  journal={arXiv preprint arXiv:1901.07042},
  year={2019}
}

@article{Johnson2019,
  year = {2019},
  publisher = {Springer Science and Business Media {LLC}},
  volume = {6},
  number = {1},
  author = {Alistair EW Johnson and Tom J Pollard and Seth J Berkowitz and Nathaniel R Greenbaum and Matthew P Lungren and Chih-ying Deng and Roger G Mark and Steven Horng},
  title = {{MIMIC}-{CXR}, a de-identified publicly available database of chest radiographs with free-text reports},
  journal = {Scientific Data}
}

@article{goldberger2000physiobank,
  title={PhysioBank, PhysioToolkit, and PhysioNet: components of a new research resource for complex physiologic signals},
  author={Goldberger, Ary L and Amaral, Luis AN and Glass, Leon and Hausdorff, Jeffrey M and Ivanov, Plamen Ch and Mark, Roger G and Mietus, Joseph E and Moody, George B and Peng, Chung-Kang and Stanley, H Eugene},
  journal={circulation},
  volume={101},
  number={23},
  pages={e215--e220},
  year={2000},
  publisher={Am Heart Assoc}
}

@inproceedings{he2016deep,
  title={Deep Residual Learning for Image Recognition},
  author={He, Kaiming and Zhang, Xiangyu and Ren, Shaoqing and Sun, Jian},
  booktitle={CVPR},
  pages={770--778},
  year={2016}
}

@inproceedings{devlin-etal-2019-bert,
    title = "{BERT}: Pre-training of Deep Bidirectional Transformers for Language Understanding",
    author = "Devlin, Jacob  and
      Chang, Ming-Wei  and
      Lee, Kenton  and
      Toutanova, Kristina",
    booktitle = "NAACL",
    year = "2019",
    publisher = "Association for Computational Linguistics",
    pages = "4171--4186",
}

@article{vaswani2017attention,
  title={Attention Is All You Need},
  author={Vaswani, Ashish and Shazeer, Noam and Parmar, Niki and Uszkoreit, Jakob and Jones, Llion and Gomez, Aidan N and Kaiser, {\L}ukasz and Polosukhin, Illia},
  journal={NIPS},
  volume={30},
  year={2017}
}

@inproceedings{papineni2002bleu,
  title={BLEU: a Method for Automatic Evaluation of Machine Translation},
  author={Papineni, Kishore and Roukos, Salim and Ward, Todd and Zhu, Wei-Jing},
  booktitle={ACL},
  pages={311--318},
  year={2002}
}

@inproceedings{lin2004rouge,
    title = "{ROUGE}: A Package for Automatic Evaluation of Summaries",
    author = "Lin, Chin-Yew",
    booktitle = "Text Summarization Branches Out",
    year = "2004",
    pages = "74--81",
}

@inproceedings{banerjee2005meteor,
  title={METEOR: An automatic metric for MT evaluation with improved correlation with human judgments},
  author={Banerjee, Satanjeev and Lavie, Alon},
  booktitle={{ACL} Workshop on Intrinsic and Extrinsic Evaluation Measures for Machine Translation and/or Summarization},
  pages={65--72},
  year={2005}
}

@inproceedings{yu2022anatomy,
  title={Anatomy-Guided Weakly-Supervised Abnormality Localization in Chest X-rays},
  author={Yu, Ke and Ghosh, Shantanu and Liu, Zhexiong and Deible, Christopher and Batmanghelich, Kayhan},
  booktitle={MICCAI},
  year={2022},
  organization={Springer}
}

@article{wang2022self,
  title={Self adaptive global-local feature enhancement for radiology report generation},
  author={Wang, Yuhao and Wang, Kai and Liu, Xiaohong and Gao, Tianrun and Zhang, Jingyue and Wang, Guangyu},
  journal={arXiv preprint arXiv:2211.11380},
  year={2022}
}

@article{yang2022knowledge,
  title={Knowledge Matters: Chest Radiology Report Generation with General and Specific Knowledge},
  author={Yang, Shuxin and Wu, Xian and Ge, Shen and Zhou, S Kevin and Xiao, Li},
  journal={MIA},
  pages={102510},
  year={2022},
  publisher={Elsevier}
}

@article{zhu2022pcan,
  title={PCAN: Pixel-wise classification and attention network for thoracic disease classification and weakly supervised localization},
  author={Zhu, Xiongfeng and Pang, Shumao and Zhang, Xiaoxuan and Huang, Junzhang and Zhao, Lei and Tang, Kai and Feng, Qianjin},
  journal={CMIG},
  volume={102},
  pages={102137},
  year={2022},
  publisher={Elsevier}
}

@article{openi,
  title={Preparing a collection of radiology examinations for distribution and retrieval},
  author={Dina Demner-Fushman and Marc D. Kohli and Marc B. Rosenman and Sonya E. Shooshan and Laritza M. Rodriguez and S. Antani and George R. Thoma and Clement J. McDonald},
  journal={JAMIA},
  year={2016},
  volume={23 2},
  pages={
          304-10
        }
}

@inproceedings{miura2021improving,
  title={Improving Factual Completeness and Consistency of Image-to-Text Radiology Report Generation},
  author={Miura, Yasuhide and Zhang, Yuhao and Tsai, Emily and Langlotz, Curtis and Jurafsky, Dan},
  booktitle={NAACL},
  pages={5288--5304},
  year={2021}
}

@article{li2021benchmarking,
  title={Benchmarking Detection Transfer Learning with Vision Transformers},
  author={Li, Yanghao and Xie, Saining and Chen, Xinlei and Dollar, Piotr and He, Kaiming and Girshick, Ross},
  journal={arXiv preprint arXiv:2111.11429},
  year={2021}
}

@inproceedings{lin2014microsoft,
  title={Microsoft COCO: common objects in context},
  author={Lin, Tsung-Yi and Maire, Michael and Belongie, Serge and Hays, James and Perona, Pietro and Ramanan, Deva and Doll{\'a}r, Piotr and Zitnick, C Lawrence},
  booktitle={ECCV},
  year={2014},
  organization={Springer}
}

@inproceedings{jain2021radgraph,
  title={RadGraph: Extracting Clinical Entities and Relations from Radiology Reports},
  author={Jain, Saahil and Agrawal, Ashwin and Saporta, Adriel and Truong, Steven and Bui, Tan and Chambon, Pierre and Zhang, Yuhao and Lungren, Matthew P and Ng, Andrew Y and Langlotz, Curtis and others},
  booktitle={NeurIPS: Datasets and Benchmarks Track (Round 1)},
  year=2021
}

@inproceedings{neumann2019scispacy,
  title={ScispaCy: Fast and Robust Models for Biomedical Natural Language Processing},
  author={Neumann, Mark and King, Daniel and Beltagy, Iz and Ammar, Waleed},
  booktitle={{BioNLP} Workshop and Shared Task},
  pages={319--327},
  year={2019}
}

@inproceedings{endo2021retrieval,
  title={Retrieval-Based Chest X-Ray Report Generation Using a Pre-trained Contrastive Language-Image Model},
  author={Endo, Mark and Krishnan, Rayan and Krishna, Viswesh and Ng, Andrew Y and Rajpurkar, Pranav},
  booktitle={MLH},
  pages={209--219},
  year={2021},
  organization={PMLR}
}

@inproceedings{syeda2020chest,
  title={Chest X-ray Report Generation through Fine-Grained Label Learning},
  author={Syeda-Mahmood, Tanveer and Wong, Ken CL and Gur, Yaniv and Wu, Joy T and Jadhav, Ashutosh and Kashyap, Satyananda and Karargyris, Alexandros and Pillai, Anup and Sharma, Arjun and Syed, Ali Bin and others},
  booktitle={MICCAI},
  pages={561--571},
  year={2020},
  organization={Springer}
}

@inproceedings{huang2017densely,
  title={Densely Connected Convolutional Networks},
  author={Huang, Gao and Liu, Zhuang and Van Der Maaten, Laurens and Weinberger, Kilian Q},
  booktitle={CVPR},
  pages={4700--4708},
  year={2017}
}

@inproceedings{tanida2023interactive,
  title={Interactive and Explainable Region-guided Radiology Report Generation},
  author={Tanida, Tim and M{\"u}ller, Philip and Kaissis, Georgios and Rueckert, Daniel},
  booktitle={CVPR},
  pages={7433--7442},
  year={2023}
}

@article{van2008visualizing,
  title={Visualizing data using t-SNE.},
  author={Van der Maaten, Laurens and Hinton, Geoffrey},
  journal={JMLR},
  volume={9},
  number={11},
  year={2008}
}

@article{dowson1982frechet,
  title={The Fr{\'e}chet distance between multivariate normal distributions},
  author={Dowson, DC and Landau, BV666017},
  journal={JMA},
  volume={12},
  number={3},
  pages={450--455},
  year={1982},
  publisher={Elsevier}
}

@book{Jolliffe,
  address = {New York},
  author = {{Jolliffe}, Ian},
  publisher = {Springer Verlag},
  title = {Principal component analysis},
  year = 2002
}

@inproceedings{smit2020combining,
  title={Combining Automatic Labelers and Expert Annotations for Accurate Radiology Report Labeling Using BERT},
  author={Smit, Akshay and Jain, Saahil and Rajpurkar, Pranav and Pareek, Anuj and Ng, Andrew Y and Lungren, Matthew},
  booktitle={EMNLP},
  pages={1500--1519},
  year={2020}
}

\appendix
\chapter*{Supplementary Material}
\section{Anatomy-Finding Faster R-CNN Architecture} \label{faster_rcnn}
\begin{figure}[h]
\centering
\includegraphics[width=0.99\textwidth]{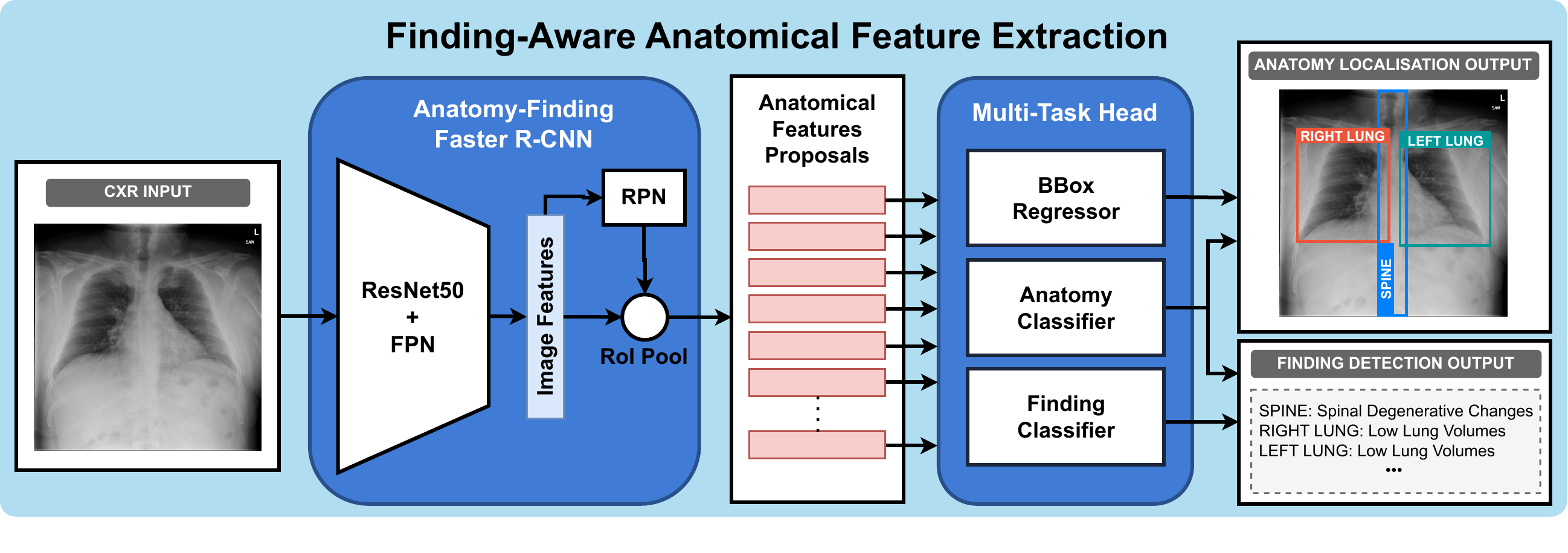}
\caption{The proposed anatomy-finding Faster R-CNN trained jointly on anatomy localisation and finding detection.} \label{fig:faster_rcnn}
\end{figure}

\section{Anatomy Localisation \& Finding Detection Results}
\label{anatfind_results}

\begin{table}[h!]
\centering
\caption{Anatomy localisation and finding detection results of different configurations of the proposed Faster R-CNN: anatomy localisation only (Anatomy); and including the finding classification head (Finding) (\textit{our proposed solution}).}
\label{tab:classification-results}
\begin{tabular}{|cc|c|c|}
\hline
\multicolumn{2}{|c|}{\textbf{Supervision}} & \textbf{Anatomy Localisation } & \textbf{ Finding Detection }\\
\textbf{ Anatomy } &  \textbf{ Finding } & \textbf{mAP@0.5} & \textbf{AUROC} \\
\hline
\checkmark & & 0.938 & - \\
\checkmark &  \checkmark & 0.918 & 0.863 \\
\hline
\end{tabular}
\end{table}

We evaluate the anatomy localisation performance of our proposed Faster R-CNN by computing the mean Average Precision (mAP@0.5), with positive detections when the Intersection over Union score between the predicted bounding boxes and the ground truth is above 0.5. Finding detection performance is measured by computing the Area Under the Receiver Operating Characteristic (AUROC) for each finding at each anatomical region. 

Table \ref{tab:classification-results} shows that our proposed anatomy-finding Faster R-CNN performs worse than a standard Faster R-CNN trained solely on anatomy localisation in terms of mAP@0.5 while achieving a good finding detection score. This is expected, as we select the best anatomy-only Faster R-CNN model based on the highest mAP@0.5 on the validation set and the anatomy-finding Faster R-CNN model based on the finding detection AUROC score. We observe that the trade-off between anatomy localisation and finding detection performance can be tuned by adjusting the weighting hyperparameter $\lambda$ in the multi-task loss.

\section{Anatomical Regions \& Findings}
\label{app:anatomy-and-finding-labels}

\begin{table}[h!]
\caption{Complete set of 36 anatomical regions and 71 findings used to supervise the anatomy localisation and the finding detection tasks, as annotated in the Chest ImaGenome dataset (\url{https://physionet.org/content/chest-imagenome/1.0.0/}).}\label{list_classes}
\resizebox{\textwidth}{!}{\begin{tabular}{llll}
\hline
\multicolumn{4}{c}{\multirow{2}{*}{\textbf{Anatomical Regions}}}                                                                                       \\
\multicolumn{4}{c}{}                                                                                                                                   \\ \hline
abdomen                     & left clavicle                            & mediastinum                           & right lower lung zone                 \\ \hline
aortic arch                 & left costophrenic angle                  & right apical zone                     & right lung                            \\ \hline
cardiac silhouette          & left hemidiaphragm                       & right atrium                          & right mid lung zone                   \\ \hline
carina                      & left hilar structures                    & right cardiac silhouette              & right upper abdomen                   \\ \hline
cavoatrial junction         & left lower lung zone                     & right cardiophrenic angle             & right upper lung zone                 \\ \hline
descending aorta            & left lung                                & right clavicle                        & spine                                 \\ \hline
left apical zone            & left mid lung zone                       & right costophrenic angle              & svc                                   \\ \hline
left cardiac silhouette     & left upper abdomen                       & right hemidiaphragm                   & trachea                               \\ \hline
left cardiophrenic angle    & left upper lung zone                     & right hilar structures                & upper mediastinum                     \\ \hline
\multicolumn{4}{c}{\multirow{2}{*}{\textbf{Findings}}}                                                                                                 \\
\multicolumn{4}{c}{}                                                                                                                                   \\ \hline
airspace opacity            & cyst/bullae                              & linear/patchy atelectasis             & pneumothorax                          \\ \hline
alveolar hemorrhage         & diaphragmatic eventration (benign)       & lobar/segmental collapse              & prosthetic valve                      \\ \hline
aortic graft/repair         & elevated hemidiaphragm                   & low lung volumes                      & pulmonary edema/hazy opacity          \\ \hline
artifact                    & endotracheal tube                        & lung cancer                           & rotated                               \\ \hline
aspiration                  & enlarged cardiac silhouette              & lung lesion                           & scoliosis                             \\ \hline
atelectasis                 & enlarged hilum                           & lung opacity                          & shoulder osteoarthritis               \\ \hline
bone lesion                 & enteric tube                             & mass/nodule (not otherwise specified) & spinal degenerative changes           \\ \hline
breast/nipple shadows       & fluid overload/heart failure             & mediastinal displacement              & spinal fracture                       \\ \hline
bronchiectasis              & goiter                                   & mediastinal drain                     & sub-diaphragmatic air                 \\ \hline
cabg grafts                 & granulomatous disease                    & mediastinal widening                  & subclavian line                       \\ \hline
calcified nodule            & hernia                                   & multiple masses/nodules               & superior mediastinal mass/enlargement \\ \hline
cardiac pacer and wires     & hydropneumothorax                        & pericardial effusion                  & swan-ganz catheter                    \\ \hline
chest port                  & hyperaeration                            & picc                                  & tortuous aorta                        \\ \hline
chest tube                  & ij line                                  & pigtail catheter                      & tracheostomy tube                     \\ \hline
clavicle fracture           & increased reticular markings/ild pattern & pleural effusion                      & vascular calcification                \\ \hline
consolidation               & infiltration                             & pleural/parenchymal scarring          & vascular congestion                   \\ \hline
copd/emphysema              & interstitial lung disease                & pneumomediastinum                     & vascular redistribution               \\ \hline
costophrenic angle blunting & intra-aortic balloon pump                & pneumonia                             &                                       \\ \hline
\end{tabular}}
\end{table}

\section{CNN Pre-Training on Chest ImaGenome Findings}
\label{cnnfind_results}

We train ResNet-101 on the set of 71 findings labels listed in Appendix \ref{app:anatomy-and-finding-labels}. Differently from Faster R-CNN, ResNet-101 is trained only to classify whether findings are present or absent in a CXR without any further anatomical localisation of each finding. We train ResNet-101 for 50 epochs, with an initial learning rate set to $10^{-3}$ and decrease every 2 epochs by a factor of $0.5$. The loss term corresponds to a binary cross-entropy:

$$\mathcal{L}_{class} = - \sum_{m=1}^M w_m f_{m} \log(p_{m}) $$

\noindent
between the predicted probability of each class $p_m$ and the ground truth $f_m$. We class weight with $w_m=(1/\nu_m)^{0.25}$, where $\nu_m$ corresponds to the frequency of the finding $f_m$ in the training set.

\vspace{30pt}

\section{Hyperparameter Search}

\begin{table}[h!]
\centering
\caption{Hyperparameter search space. We highlight in \textbf{bold} the chosen values.}
\label{tab:hyperparameters}
\begin{tabular}{|l|c|c|}
\hline
\textbf{Model} & \textbf{Hyperparameter} & \textbf{Search Space} \\
\hline
\hline
\multicolumn{1}{|c|}{\multirow{3}{3cm}{\textbf{Anatomy-Finding \\ Faster R-CNN}}} & learning rate & \{0.001, 0.0005, \textbf{0.0001}, 0.00005\} \\
& $\lambda$ & \{1, 10, \textbf{100}, 1000\} \\
& $\alpha$ & \{0, \textbf{0.25}, 0.50, 0.75, 1\} \\
\hline
\hline
\multirow{1}{3cm}{{\textbf{Triples Extractor}}} &
learning rate & \{0.0005, \textbf{0.0001}, 0.00005\} \\
\hline
\hline
\multirow{2}{3cm}{{\textbf{Report Generator}}} & learning rate & \{0.0005, \textbf{0.0001}, 0.00005\} \\
& triples masking \% & \{30, 40, \textbf{50}, 60, 70\} \\
\hline
\end{tabular}
\end{table}

\newpage
\section{Example predicted reports}\label{examples}

\begin{figure}[h!]
\includegraphics[width=\textwidth]{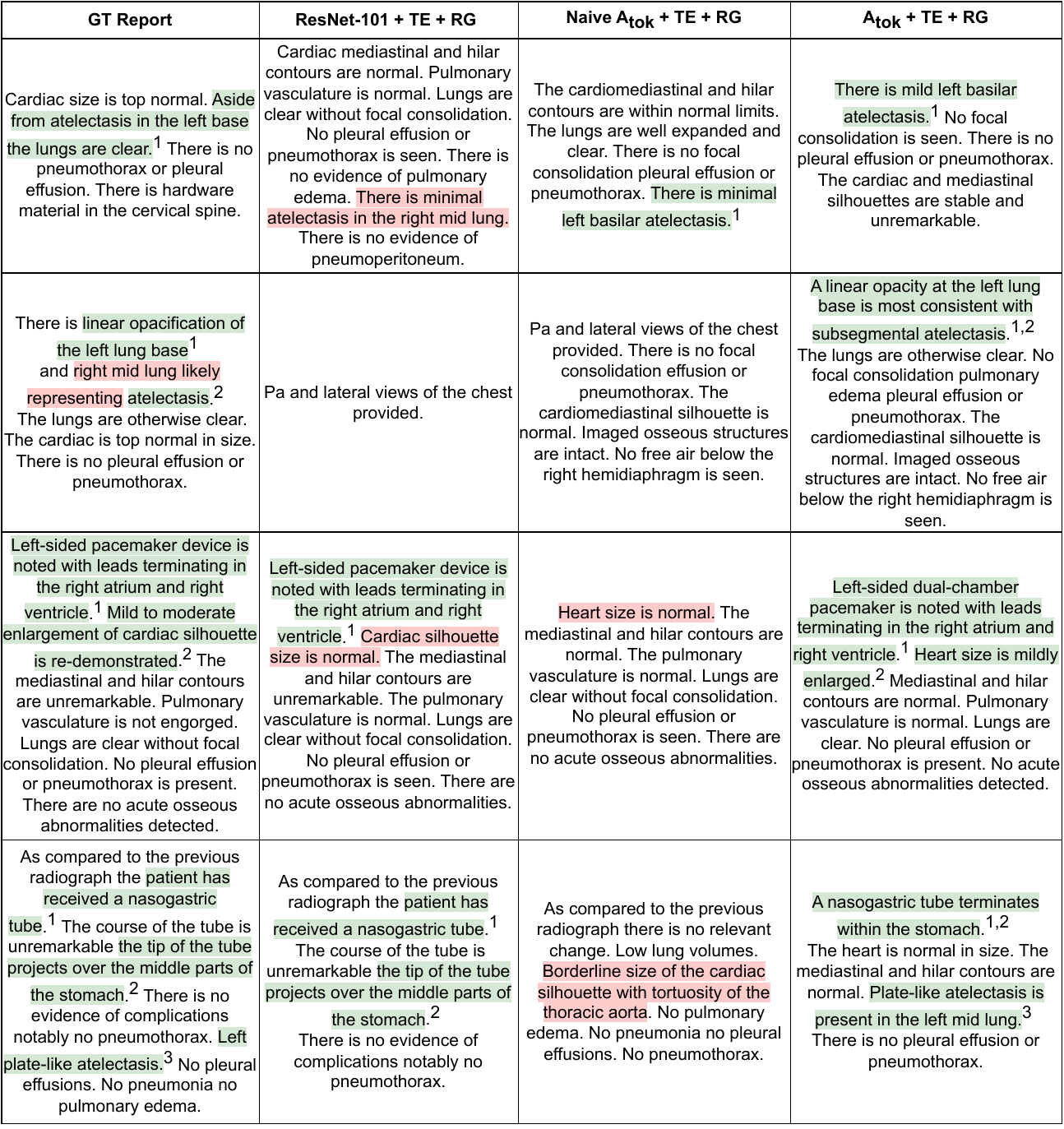}
\caption{Examples of predicted reports with different visual representations (examples selected randomly from the subset of reports where predictions were different between representation methods). From left to right: the ground truth (GT) report, the predicted reports using a CNN, the naive anatomical tokens and the finding-aware anatomical tokens as the visual representations. In generated reports, correctly detected positive findings are highlighted in green, and errors are highlighted in red. The equivalent text spans in the ground truth report are also highlighted; we number corresponding descriptions.}
\end{figure}

\end{document}